%% file: main.tex
\documentclass[10pt,twocolumn,letterpaper]{article}

\usepackage[pagenumbers]{cvpr} 

\input{predefine}
\input{preamble}

\newcommand{\ours}{{\texttt{LumiX}}\xspace}

\definecolor{cvprblue}{rgb}{0.21,0.49,0.74}
\usepackage[pagebackref,breaklinks,colorlinks,allcolors=cvprblue]{hyperref}

\def\paperID{1} 
\def\confName{CVPR}
\def\confYear{2026}

\title{\ours: Structured and Coherent Text-to-Intrinsic Generation}

\author{Xu Han$^{1,2\diamond}$ \qquad Biao Zhang$^{2}$ \qquad Xiangjun Tang$^{2}$ \qquad Xianzhi Li$^{1}$$^\dag$ \qquad Peter Wonka$^{2}$$^\dag$\\
$^1$HUST \qquad $^2$KAUST\\
}

\begin{document}
\maketitle

\renewcommand{\thefootnote}{}
\footnotetext{$^\dag$ Corresponding authors.}
\footnotetext{$^\diamond$ This work was done during Xu Han’s internship at KAUST.}

\input{pic/teaser}

\input{sec/0_abstract}    
\input{sec/1_intro}

\input{sec/2_related_work}

\input{sec/3_method}

\input{sec/4_experiment}
\input{sec/5_conclusion}
{
    \small
    \bibliographystyle{ieeenat_fullname}
    \bibliography{main}
}

\input{sec/X_suppl}

\end{document}

%% file: predefine.tex
\usepackage{graphicx}
\usepackage{amsmath}
\usepackage{amssymb}
\usepackage{booktabs}
\usepackage{blindtext}
\usepackage{bbding}

\definecolor{pretty-blue}{RGB}{0, 113, 188}
\definecolor{linecolor}{gray}{.89} 
\definecolor{linecolor3}{gray}{.95} 
\definecolor{linecolor2}{gray}{.98} 
\definecolor{linecolor1}{gray}{.95} 

\definecolor{pretraincolor}{HTML}{2E59A7} 

\definecolor{line3}{HTML}{F0FEF0}
\definecolor{line2}{HTML}{DCFCF9}
\definecolor{line1}{HTML}{B3F7F4}

\definecolor{mambacolor}{HTML}{6424D6}
\definecolor{reconcolor}{HTML}{412F8A}
\definecolor{resnetcolor}{HTML}{8DA0CB}
\definecolor{vitcolor}{HTML}{fc8e62}

\definecolor{newgreen}{HTML}{019C74}
\definecolor{berryred}{HTML}{ED656B}
\definecolor{chinablue}{HTML}{66A9C9}

\usepackage{times}
\usepackage{epsfig}

\usepackage{multirow}
\usepackage{bbm}
\usepackage{multirow}
\usepackage{algorithm}
\usepackage[noend]{algpseudocode}
\usepackage{tabularx}
\usepackage{makecell}
\usepackage{microtype}
\usepackage{textcomp}
\usepackage{colortbl}
\usepackage{url}
\usepackage{cuted}

\usepackage{caption}
\usepackage[outline]{contour}

\usepackage{pifont}  

\definecolor{zimabule}{HTML}{272CE5}
\definecolor{bulered}{HTML}{6527fe}
\definecolor{greenblue}{HTML}{00acc1}
\definecolor{orangered}{HTML}{ed635e}
\definecolor{orange}{HTML}{ff4e00} 
\definecolor{lightred}{HTML}{ff3886} 
\definecolor{lightpurple}{HTML}{F0D1F7} 

\definecolor{method1}{HTML}{c2c7fb}
\definecolor{method2}{HTML}{a0a7fa}
\definecolor{method3}{HTML}{7681f8}
\definecolor{method4}{HTML}{535fe8}
\definecolor{method5}{HTML}{3a44b0}

\definecolor{figure6}{HTML}{CE2C6A}

\usepackage{wrapfig} 
\usepackage{soul}

\usepackage{tikz}
\usepackage{xparse}

\NewDocumentCommand{\Circled}{ O{} m }{%
  \tikz[baseline=(char.base)]{
    \node[
      circle,
      draw,
      inner sep=1pt,
      minimum size=2.1ex,
      line width=0.35pt,
      font=\sffamily\footnotesize,   
      #1
    ] (char) {#2};
  }%
}

\usepackage{threeparttable}

%% file: preamble.tex
\usepackage{comment}

%% file: pic/teaser.tex
\begin{strip}
\begin{minipage}{\textwidth}\centering
\vspace{-5em}
\includegraphics[width=1.0\textwidth]{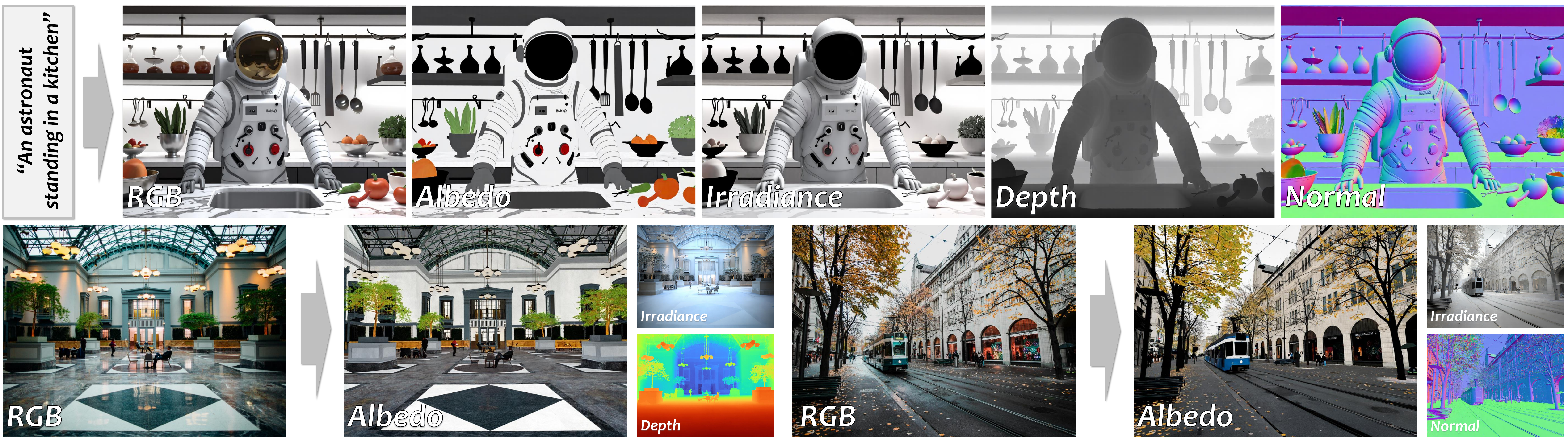}
\vspace{-15pt}
\captionof{figure}{
\textbf{\ours for Text-to-Intrinsic Generation.}
Given a text prompt, \ours jointly generates a coherent set of intrinsic maps, including RGB color, albedo, irradiance, depth, and normal.  
Built on a powerful diffusion prior, it produces diverse and physically grounded intrinsic images, and can also perform image-conditioned intrinsic decomposition even though it is trained with text-only conditioning.
}
\label{fig:teaser}
\end{minipage}
\end{strip}

%% file: sec/0_abstract.tex
\begin{abstract}
We present \ours, a structured diffusion framework for coherent text-to-intrinsic generation.  
Conditioned on text prompts, \ours jointly generates a comprehensive set of intrinsic maps (\eg, albedo, irradiance, normal, depth, and final color), providing a structured and physically consistent description of an underlying scene.  
This is enabled by two key contributions:
1) Query-Broadcast Attention, a mechanism that ensures structural consistency by sharing queries across all maps in each self-attention block.
2) Tensor LoRA, a tensor-based adaptation that parameter-efficiently models cross-map relations for efficient joint training.
Together, these designs enable stable joint diffusion training and unified generation of multiple intrinsic properties.  
Experiments show that \ours produces coherent and physically meaningful results, achieving {23\% higher alignment} and a {better preference score} (0.19 vs.\ -0.41) compared to the state of the art, and it can also perform image-conditioned intrinsic decomposition within the same framework.

\end{abstract}

\input{pic/Method_Overview}

%% file: pic/Method_Overview.tex
\begin{figure*}[t]
    \begin{center}
    \vspace{-4pt}
    \includegraphics[width=1.0\linewidth]{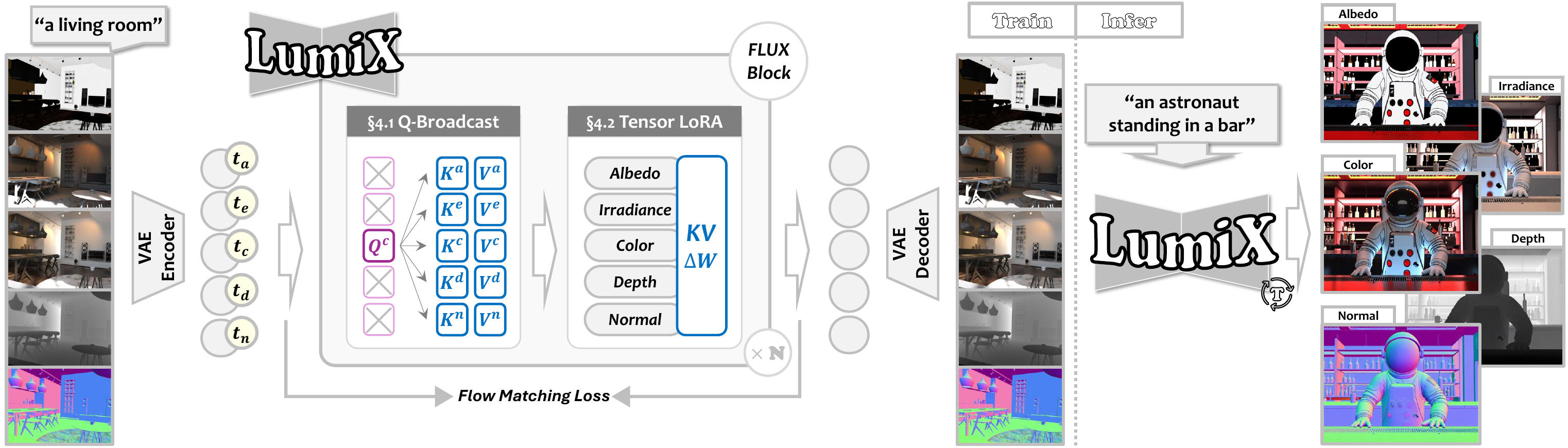}
    \vspace{-15pt}
\caption{
\textbf{Overview of \ours.} Our goal is to generate a coherent set of intrinsic maps from text.  
\textbf{Left: Training.} Multiple intrinsic images are encoded into the latent space and concatenated along the batch dimension.  
We introduce \emph{Query-Broadcast Attention} (\cref{sec:cross-amp-attn}) to ensure pixel alignment across properties, 
and \emph{Tensor LoRA} (\cref{sec:tensor-lora}) to efficiently finetune the $KV$ projections for each property.  
Different timesteps are assigned to different properties for flexible conditioning.  
\textbf{Right: Inference.} Given a text or image input, \ours jointly outputs all intrinsic maps in a single forward pass, 
supporting both text-to-intrinsic generation and intrinsic decomposition.
}
    \label{fig:method_overview}
    \vspace{-20pt}
    \end{center}
\end{figure*}

%% file: sec/1_intro.tex
\section{Introduction}
\label{sec:intro}

Recent advances in text-to-image diffusion models~\cite{rombach2022high} have made it possible to generate realistic and detailed images from natural language.  
However, these models still produce a single RGB image and do not reveal the underlying structure of the scene, such as geometry, lighting, or material properties.  
For many vision and graphics tasks, a single image is not enough.  
We often need a structured representation that separates these intrinsic factors, including albedo, irradiance, normal, depth, and the final color.  
Together, these maps describe both the physical and semantic aspects of a scene.  
This raises an important question: can we directly generate a coherent set of intrinsic maps from text?  
In this work, we study this problem under the term \emph{text-to-intrinsic generation} and aim to build a unified model that can jointly generate scene content and physical attributes.

Most existing studies focus on {intrinsic image decomposition}~\cite{grosse2009ground,chen2013simple,li2018learning,careaga2023intrinsic}, where a model receives a rendered or captured image and predicts its intrinsic components such as albedo, shading, or normal.  
This formulation is effective for analysis but inherently limited, as it depends on a given image and cannot generate new scenes from language.  
\emph{Text-to-intrinsic generation} removes this restriction.  
It starts from text and produces both the scene content and its physical attributes within a single generative process.  
The same diffusion backbone can also switch to image-conditioned decomposition during inference, enabling both generation and understanding within one unified framework.  
Therefore, text-to-intrinsic generation is not merely a generalization of intrinsic decomposition, but a broader step toward unified multi-attribute generation.

The main challenge in {text-to-intrinsic generation} is maintaining cross-map structural consistency under text-only conditioning.  
In image-conditioned settings such as intrinsic decomposition or RGB-to-X translation~\cite{careaga2024colorful, zeng2024rgb}, all outputs share the same input image, so spatial alignment is naturally preserved.  
In contrast, each intrinsic map here is generated from its own noise sample and is only weakly coupled through the shared text embedding.  
Without an explicit spatial anchor, structures can drift across maps: an object may appear in the color map but disappear in the normal map, or the geometry in the depth map may not match the shading in the irradiance map.  
Ensuring consistent scene structure across all intrinsic maps without image supervision is therefore the central problem of this task.

To address the consistency challenge across multiple images, related works have explored two strategies.  
One strategy~\cite{byung2025jointdit} trains each intrinsic map independently, using separate models or LoRA~\cite{hu2022lora} branches.  
With strong diffusion priors, these models can produce high-quality albedo, normal, or depth maps, but the results often lack semantic alignment across maps.  
Another strategy~\cite{kocsis2025intrinsix} builds a fully coupled framework that concatenates features from all maps and applies cross-intrinsic attention to enforce global interaction.  
This improves consistency but makes training unstable, and computation scales quadratically with the number of maps.  
Overall, these directions reveal a clear trade-off between consistency and efficiency, motivating a more structured sharing mechanism that maintains semantic coherence while preserving property-specific independence.

We present a systematic analysis of the pipeline, examining it from two critical perspectives: the \textit{forward pass} and the \textit{training/finetuning} process. In the forward pass, we ensure structural consistency by broadcasting identical queries to all intrinsic maps (a mechanism we term \textbf{Query-Broadcast Attention}). This design is grounded in prior research on content and style separation~\cite{chung2024style, hertz2024style}, which establishes that queries primarily capture content, whereas keys and values contribute to style in self-attentions. For the training/finetuning process, we present several possible low-rank-based (LoRA) finetuning strategies. We extend the conventional matrix-based LoRA to a tensor-based formulation. We demonstrate that our proposed \textbf{Tensor LoRA} not only yields the best quality results but also effectively preserves structural consistency and operates with high efficiency.
Our main contributions:
\begin{itemize}
    \item We present {\ours}, a structured diffusion framework for coherent text-to-intrinsic generation that jointly produces multiple physically consistent intrinsic maps.  
    \item We design {Query-Broadcast Attention} to align scene content across maps by sharing queries while keeping attribute projections independent.  
    \item We propose {Tensor LoRA}, a tensor-based low-rank adaptation that models global cross-map relations in a parameter-efficient and scalable way.
\end{itemize}

%% file: sec/2_related_work.tex
\section{Related Work}
\label{sec:related_work}

\subsection{Intrinsic and Multi-map Generation}
Intrinsic image decomposition has long focused on separating an image into physically meaningful factors such as albedo, shading, and materials~\cite{land1971lightness,chen2013simple,janner2017self,li2018learning,sengupta2019neural,wang2021learning,careaga2023intrinsic,careaga2024colorful}.  
Most methods are image-conditioned and rely on predefined physical priors, such as albedo–shading consistency or inverse–forward rendering cycles~\cite{liang2025diffusionrenderer,chen2024intrinsicanything,zheng2025dnf,sun2025ouroboros,chen2025uni}.  
While effective for analysis, they cannot synthesize new scenes or handle open-vocabulary text input.
To move beyond decomposition, later works predict multiple complementary maps such as depth~\cite{ke2024repurposing}, normal~\cite{ye2024stablenormal}, or segmentation~\cite{zhao2025diception}. 
Feed-forward~\cite{zhu2022irisformer,zhu2022learning} and GAN-based models~\cite{bhattad2023stylegan} support multi-map prediction but require dense supervision and often lose structural coherence.

Diffusion-based methods~\cite{kocsis2024intrinsic,luo2024intrinsicdiffusion,zeng2024rgb} extend text-to-image backbones or add task tokens for multi-map generation, but sharing one backbone across maps easily leads to interference as the number of maps grows.  
Recent designs add per-map LoRA adapters~\cite{lopes2024material} or communication layers~\cite{vainer2024collaborative,byung2025jointdit}, though they do not clearly separate shared structure from property-specific variation.
More recently, text-to-X diffusion models~\cite{dirik2025prism,xue2025diffusion,kocsis2025intrinsix} explore intrinsic generation without image supervision.  
Among them, 
IntrinsiX~\cite{kocsis2025intrinsix} improves consistency with a single fused attention space, but this limits scalability and interpretability.
In contrast, our method adopts a structured formulation that preserves pixel alignment across intrinsic maps without heavy fusion, enabling scalable and coherent text-to-intrinsic generation.

\subsection{Cross-map Attention and Adaptation}
Attention layers are central to diffusion models, as they integrate conditional information.
Many works manipulate the query, key, or value projections to control generation~\cite{chefer2023attend,zhang2023jointnet,wei2023elite,cao2023masactrl}.
Queries typically capture spatial and semantic content, while keys and values encode appearance or modality cues~\cite{hertz2024style,ye2023ip}.
Attend-and-Excite~\cite{chefer2023attend} modifies Q to highlight object presence, and MasaCtrl~\cite{cao2023masactrl} mixes Q, K, and V across images to improve spatial consistency. Other methods introduce separate K/V branches~\cite{zhang2023adding,zhang2023jointnet} or adapter modules~\cite{xie2024styletex,ye2023ip,kumari2023multi} to inject conditions such as segmentation, depth, or reference images~\cite{alaluf2024cross,hertz2024style}. Recent works extend these ideas to multi-map diffusion by aligning K/V projections across maps or by learning property-specific branches~\cite{byung2025jointdit,zhou2025attention}.
These methods focus on external conditioning or reference, rather than ensuring structural consistency between maps generated from text.
Our Query-Broadcast Attention reframes attention for intrinsic map alignment by sharing queries across maps while keeping keys and values independent, allowing shared scene structure and property-specific variation to be preserved.

\subsection{Parameter-efficient and Tensorized Adaptation}
Low-rank adaptation methods such as LoRA~\cite{hu2022lora,zhang2023adaptive,liu2024dora,kopiczko2024vera} fine-tune large models efficiently by adding small low-rank updates to frozen weights~\cite{han2024parameterefficient}.
In multi-map generation, assigning an independent LoRA to each map~\cite{lopes2024material,byung2025jointdit,vainer2024collaborative} ignores cross-map relations and causes quadratic parameter growth as the number of maps increases.  
To improve efficiency, tensor decomposition methods such as tensor-train~\cite{oseledets2011tensor} and tensor-ring~\cite{zhao2016tensor} provide a compact way to represent high-dimensional parameters.
Following this idea, we design Tensor LoRA, which encodes all LoRA updates in one structured tensor and factorizes it into shared cores and per-map components.
This formulation captures cross-map relations while keeping the parameter cost close to linear.

%% file: sec/3_method.tex
\section{Preliminaries}
\label{sec:prelim}

\paragraph{Finetuning with LoRA.} In a denoising diffusion network, each self-attention layer accepts a hidden feature $H\in\mathbb{R}^{d\times L}$ as input,
\begin{equation}
    \mathrm{SA}(H) = \mathrm{softmax}\left(
        QK^\intercal/\sqrt{d}
    \right)V,
\end{equation}
where $Q=W_QH$, $K=W_KH$, and $V=W_VH$. The learnable parameters are $\{W_Q\in\mathbb{R}^{d\times d}, W_K\in\mathbb{R}^{d\times d}, W_V\in\mathbb{R}^{d\times d}\}$. Low-Rank Adaptation (LoRA) is applied for efficient finetuning,
\begin{equation}\label{eq:lora}
    W \leftarrow W + \Delta, \quad \Delta = AB^\intercal,
\end{equation}
where $A\in\mathbb{R}^{d\times R}$, $B\in\mathbb{R}^{d\times R}$ are learnable low-rank adapters. Usually, $R\ll d$. For a single token $h\in\mathbb{R}^d$, the calculation is often done by
\begin{equation}\label{eq:update-lora}
    Wh+A(B^\intercal h),
\end{equation}
which is a more efficient version of $(W+AB^\intercal)h$.

\paragraph{Flow matching training.} To train flow matching on a dataset $\mathcal{D}$, we minimize the loss function,
\begin{equation}
    \min_\theta \mathbb{E}_{t\in[0, 1], \boldsymbol{\epsilon}\sim\mathcal{N}(0, 1),\mathbf{z}\sim\mathcal{D}}\left\|v_\theta(\mathbf{z}_t,t,\mathcal{C}) - (\boldsymbol{\epsilon} - \mathbf{z})\right\|,
\end{equation}
where $\mathcal{C}$ denotes the conditioning input, and $v_\theta$ is the denoising network parameterized by $\theta$. Most models, such as Stable Diffusion~\cite{rombach2022high} and FLUX~\cite{flux2023}, are trained in the latent space $\mathbf{z}$ instead of the original pixel space $\mathbf{x}$.

\section{Method}
Given a text prompt condition $\mathcal{C}$, our goal is to jointly generate pixel-aligned intrinsic maps of the same scene: color $\mathbf{x}^{(c)}$, albedo $\mathbf{x}^{(a)}$, irradiance $\mathbf{x}^{(i)}$, depth $\mathbf{x}^{(d)}$, and normal $\mathbf{x}^{(n)}$. We denote their corresponding latent representations, obtained via a pretrained autoencoder: $\mathbf{z}^{(c)}$, $\mathbf{z}^{(a)}$, $\mathbf{z}^{(i)}$, $\mathbf{z}^{(d)}$, and $\mathbf{z}^{(n)}$, respectively. 

These maps share a common spatial layout but capture different intrinsic properties, including appearance, reflectance, illumination, and geometry. 
Our task has two primary objectives: (1) \textbf{consistency}: defined as structural consistency across properties to ensure shared content and pixel alignment, and (2) \textbf{quality}: defined as realism within each property to preserve its distinct characteristics.

We address this task by finetuning existing text-to-image diffusion models. As a starting point for our discussion, we consider the naive solution to train $M=5$ independent models, one for each map (color, albedo, irradiance, depth, and normal). This approach promises good per-map \emph{quality}, but fails to maintain structural \emph{consistency} across the maps, as they are generated in isolation.

To ensure \emph{consistency}, we introduce two core components that link the different maps together: Query-Broadcast Attention in~\cref{sec:cross-amp-attn} and a Tensor LoRA in~\cref{sec:tensor-lora}. The former modifies the forward pass to enable information exchange across all maps, while the latter improves training efficiency by proposing a lightweight finetuning strategy.

\subsection{Forward with Query-Broadcast Attention}\label{sec:cross-amp-attn}

A baseline approach is to compute the self-attention for each model $m\in\mathcal{M}=\{c, a, i, d, n\}$ (color, albedo, irradiance, depth, and normal) independently,
\begin{equation}\label{eq:vanilla-attn}
    H^{(m)}\leftarrow \mathrm{softmax}\left(Q^{(m)}K^{(m)\intercal}/\sqrt{d}\right)V^{(m)}.
\end{equation}
The design, hereafter referred to as \textbf{Vanilla Attention}, evidently ignores potential interactions between the different models in $\mathcal{M}$. 

IntrinsiX~\cite{kocsis2025intrinsix} proposed concatenating $K$ and $V$ from all models,
\begin{equation}\label{eq:cross-int-attn}
    H^{(m)} =\mathrm{softmax}\left(
        Q^{(m)}
        \textcolor{blue}{
            K^{(\mathcal{M})\intercal}
        }
        /\sqrt{d}
    \right)
    \textcolor{blue}{
        V^{(\mathcal{M})}
    },
\end{equation}
where $K^{(\mathcal{M})}=\mathrm{Concat}([K^{(c)}, K^{(a)}, \cdots])$, $V^{(\mathcal{M})}=\mathrm{Concat}([V^{(c)}, V^{(a)}, \cdots])$. This approach, termed \textbf{Cross-Intrinsic Attention} in~\cite{kocsis2025intrinsix}, improves consistency by exchanging information between models. However, its computational complexity is $M$ times larger than the vanilla approach~(\cref{eq:vanilla-attn}).

To address this issue, we propose a more efficient design.
Our approach is inspired by~\cite{hertz2024style,chung2024style,zhou2025attention}, which argue that the $K$ and $V$ matrices capture modality-specific \textit{style} (e.g., reflectance, illumination, or surface-normal variations), and that the $Q$ matrix encodes the scene's \textit{content}.
Based on this insight, we broadcast the color model's query matrix (\ie, $Q^{(c)}$) to all models.
Specifically, for each model $m$,
\begin{equation}\label{eq:broadcast-query-attn}
\boxed{
    H^{(m)}\leftarrow \mathrm{softmax}\left(
    \textcolor{blue}{
        Q^{(c)}
    }
    K^{(m)\intercal}/\sqrt{d}
    \right)
    V^{(m)}
}
\end{equation}
We refer to the design as \textbf{Query-Broadcast Attention}.
In this formulation, all models utilize the shared query $Q^{(c)}$ from the color map model.
We empirically demonstrate that our strategy in~\cref{eq:broadcast-query-attn} outperforms IntrinsiX in~\cref{eq:cross-int-attn} in later sections, while being significantly more efficient.

\subsection{Finetuning with Tensor LoRA}\label{sec:tensor-lora}
As discussed above, $Q^{(c)}$ is broadcast to all models. Consequently, we do not fine-tune the query projection $W_Q$.
The learnable parameters for a single attention layer are
\begin{equation}\label{eq:learnable-param}
    \left\{(W_K^{(c)}, W_V^{(c)}), (W_K^{(a)}, W_V^{(a)}), (W_K^{(i)}, W_V^{(i)}), \cdots\right\},
\end{equation}
which results in $2 \times M\times d\times d$ parameters in total.
When finetuning with LoRA, we do not directly update the original weight matrices $W$. Instead, we only consider the low-rank updates,
\begin{equation}
    \underbrace{
        Wh
    }_{\text{base model}}
    +
    \underbrace{
        AB^\intercal h
    }_{\text{LoRA update }}
\end{equation}
In the following, we discuss different designs of this LoRA update.

\input{tab/Method_Ablation_Comparison}

\paragraph{\Circled[fill=method1,draw=method1,text=white]{\textbf{S}} Separate LoRA.}
For a single weight matrix $W^{(m)}$ and its update $\Delta^{(m)}$, we apply LoRA in~\cref{eq:lora} to obtain the parameters $A^{(m)}\in\mathbb{R}^{d\times R}$ and $B^{(m)}\in\mathbb{R}^{d\times R}$. The resulting update for each map $m$'s activation is calculated as $A^{(m)}(B^{(m)\intercal}h^{(m)})$.
While this approach is efficient, it ignores any interaction between the $M$ models, and thus fails to maintain structural consistency. This independent update process can be visualized in a compact block-diagonal form for all $M$ models,
\begin{equation}\label{eq:separate-lora}
    \begin{bmatrix}
        \Delta^{(c)} & 0 & 0 & 0 & 0 \\
        0 & \Delta^{(a)} & 0 & 0 & 0 \\
        0 & 0 & \Delta^{(i)} & 0 & 0 \\
        0 & 0 & 0 & \Delta^{(d)} & 0 \\
        0 & 0 & 0 & 0 & \Delta^{(n)} \\
    \end{bmatrix}
    \begin{bmatrix}
        h^{(c)} \\
        h^{(a)} \\
        h^{(i)} \\
        h^{(d)} \\
        h^{(n)} \\
    \end{bmatrix}
\end{equation}

\paragraph{\Circled[fill=method2,draw=method2,text=white]{\textbf{F}} Fused LoRA.}
A straightforward solution is to fuse all the activations $h^{(m)}\in\mathbb{R}^d$ from all models into a single, concatenated vector $h^{(\mathcal{M})}\in\mathbb{R}^{Md}$.
A single LoRA update is then applied on a large update matrix $\Delta^{(\mathcal{M})}\in\mathbb{R}^{Md\times Md}$, which is further decomposed into low-rank matrices $A^{(\mathcal{M})}\in\mathbb{R}^{Md\times R}$ and $B^{(\mathcal{M})}\in\mathbb{R}^{Md\times R}$. The fused output $A^{(\mathcal{M})}(B^{(\mathcal{M})\intercal}h^{(\mathcal{M})})$, is subsequently split back into $M$ separate models. Unlike \textbf{Separate LoRA}, this approach creates a dense update matrix where the off-diagonal blocks are non-zero, explicitly modeling the interactions between all models:

\begin{equation}\label{eq:fused-lora}
    \underbrace{
    \begin{bmatrix}
        \Delta^{(cc)} & \Delta^{(ca)} & \Delta^{(ci)} & \Delta^{(cd)} & \Delta^{(cn)} \\
        \Delta^{(ac)} & \Delta^{(aa)} & \Delta^{(ai)} & \Delta^{(ad)} & \Delta^{(an)} \\
        \Delta^{(ic)} & \Delta^{(ia)} & \Delta^{(ii)} & \Delta^{(id)} & \Delta^{(in)} \\
        \Delta^{(dc)} & \Delta^{(da)} & \Delta^{(di)} & \Delta^{(dd)} & \Delta^{(dn)} \\
        \Delta^{(nc)} & \Delta^{(na)} & \Delta^{(ni)} & \Delta^{(nd)} & \Delta^{(nn)} \\
    \end{bmatrix}
    }_{\Delta^{(\mathcal{M})}\in\mathbb{R}^{Md\times Md}}
    \underbrace{
    \begin{bmatrix}
        h^{(c)} \\
        h^{(a)} \\
        h^{(i)} \\
        h^{(d)} \\
        h^{(n)} \\
    \end{bmatrix}
    }_{h^{(\mathcal{M})}\in\mathbb{R}^{Md}}
\end{equation}

\definecolor{myhigh}{RGB}{0, 114, 178}
\definecolor{mylow}{RGB}{213, 94, 0}
\paragraph{\Circled[fill=method3,draw=method3,text=white]{\textbf{H}} Hybrid LoRA.}
Inspired by the block-matrix form of \textbf{Separate LoRA} in~\cref{eq:separate-lora} and \textbf{Fused LoRA} in~\cref{eq:fused-lora}, we propose a hybrid approach.
We enhance \textbf{Separate LoRA} with non-zero off-diagonal entries. Specifically, we use higher rank ($R_1$) decomposition in \textcolor{myhigh}{diagonal} entries and lower rank ($R_2 < R_1$) in \textcolor{mylow}{off-diagonal} entries,

\begin{equation}\label{eq:hybrid-lora}
    \begin{bmatrix}
        \textcolor{myhigh}{\Delta^{(cc)}} & \textcolor{mylow}{\Delta^{(ca)}} & \textcolor{mylow}{\Delta^{(ci)}} & \textcolor{mylow}{\Delta^{(cd)}} & \textcolor{mylow}{\Delta^{(cn)}} \\
        \textcolor{mylow}{\Delta^{(ac)}} & \textcolor{myhigh}{\Delta^{(aa)}} & \textcolor{mylow}{\Delta^{(ai)}} & \textcolor{mylow}{\Delta^{(ad)}} & \textcolor{mylow}{\Delta^{(an)}} \\
        \textcolor{mylow}{\Delta^{(ic)}} & \textcolor{mylow}{\Delta^{(ia)}} & \textcolor{myhigh}{\Delta^{(ii)}} & \textcolor{mylow}{\Delta^{(id)}} & \textcolor{mylow}{\Delta^{(in)}} \\
        \textcolor{mylow}{\Delta^{(dc)}} & \textcolor{mylow}{\Delta^{(da)}} & \textcolor{mylow}{\Delta^{(di)}} & \textcolor{myhigh}{\Delta^{(dd)}} & \textcolor{mylow}{\Delta^{(dn)}} \\
        \textcolor{mylow}{\Delta^{(nc)}} & \textcolor{mylow}{\Delta^{(na)}} & \textcolor{mylow}{\Delta^{(ni)}} & \textcolor{mylow}{\Delta^{(nd)}} & \textcolor{myhigh}{\Delta^{(nn)}} \\
    \end{bmatrix}
    \begin{bmatrix}
        h^{(c)} \\
        h^{(a)} \\
        h^{(i)} \\
        h^{(d)} \\
        h^{(n)} \\
    \end{bmatrix}
\end{equation}

\paragraph{\Circled[fill=method4,draw=method4,text=white]{\textbf{T}} Tensor LoRA.} 
In our experiments, we found that both \textbf{Fused LoRA} and \textbf{Hybrid LoRA} exhibit better consistency but suffer from inefficiency due to the large dimensionality of their update matrices.
Inspired by the element-wise form of LoRA ($\Delta=AB^\intercal$),
\begin{equation}
   \Delta[i,j] =\sum^r_{\alpha=1}A[i,\alpha]B[j,\alpha],
\end{equation}
we propose to represent $\Delta^{(\mathcal{M})}\in\mathbb{R}^{Md\times Md}$ using a tensor decomposition. Specifically, we reshape the matrix into a 4th-order tensor 
$\Delta^{(\mathcal{M})}\in\mathbb{R}^{N\times d_{\text{out}}\times M \times d_{\text{in}}}$, where $M$ is the number of input models and $N$ is the number of output models.  As a general case, we allow $M\neq N$ and $d_{\text{in}}\neq d_{\text{out}}$.
The element-wise form of our proposed decomposition is
\begin{equation}\label{equ:tensor_lora}
\boxed{
\begin{aligned}
    &\Delta^{(\mathcal{M})}[i,j,k,l]\\
    =&\sum^{R_1}_{\alpha_1=1}\sum^{R_2}_{\alpha_2=1} A[i, j, \alpha_1]B[i, k, \alpha_2]C[i, l, \alpha_1, \alpha_2],
\end{aligned}
}
\end{equation}
where $A\in\mathbb{R}^{N\times d_{\text{out}} \times R_1}$, $B\in\mathbb{R}^{N\times M \times R_2}$, $C\in\mathbb{R}^{N\times d_{\text{in}} \times R_1 \times R_2}$. This is related to tensor-train (TT) decompositions~\cite{oseledets2011tensor}. 
The output is calculated using tensor contractions (\textsf{einsum} is a built-in function in many libraries such as NumPy, PyTorch and Jax):
\begin{verbatim}
# N d R1 R2, M d -> N M R1 R2
Ch = einsum('ndrs, md -> nmrs', C, h)
# N M R2, N M R1 R2 -> N R1
BCh = einsum('nms, nmrs -> nr', B, Ch)
# N d R1, N R1 -> N d
ABCh = einsum('ndr, nr -> nd', A, BCh)
\end{verbatim}
In practice, we set $M=N$, $d_{\text{in}}=d_{\text{out}}$, and $R_1 = R_2$.

\input{pic/Comparison_Results}

%% file: tab/Method_Ablation_Comparison.tex
\begin{table*}[t!]
\centering
\renewcommand{\arraystretch}{1.2}
\caption{
\textbf{Comparison of attention and LoRA designs for text-to-intrinsic generation on our Hypersim test set}~\cite{roberts2021hypersim}.  
\#P: trainable parameters (M) per attention block;  
FLOPs: FLOPs (G) of LoRA and attention per block;  
Align.: cross-map consistency;  
Other columns: human preference scores~\cite{xu2023imagereward,kirstain2023pick} across intrinsic maps.  
\textsuperscript{$\dagger$}Without the first training stage in IntrinsiX~\cite{kocsis2025intrinsix}.  
\textsuperscript{$\ddagger$}Official IntrinsiX repository.
}
\label{tab:component_ablation}

\resizebox{\linewidth}{!}{
\begin{tabular}{lcccccccccccc}
\toprule[0.7pt]
\multirow{2}{*}{\textbf{Method}} &
\multirow{2}{*}{\texttt{LoRA}} &
\multirow{2}{*}{\textbf{\#P$\downarrow$}} &
\multicolumn{1}{c}{\textbf{FLOPs$\downarrow$}} &
\multirow{2}{*}{\textbf{Align.$\uparrow$}} &
\multicolumn{2}{c}{\textbf{Color}$\uparrow$} &
\multicolumn{2}{c}{\textbf{Albedo}$\uparrow$} &
\multicolumn{2}{c}{\textbf{Irradiance}$\uparrow$} &
\multicolumn{1}{c}{\textbf{Avg.$\uparrow$}} \\
\cmidrule(lr){4-4} \cmidrule(lr){6-7} \cmidrule(lr){8-9} \cmidrule(lr){10-11} \cmidrule(lr){12-12}
& & & \texttt{LoRA}\,/\,\texttt{Attn} & &
\textbf{ImageReward} & \textbf{PickScore} &
\textbf{ImageReward} & \textbf{PickScore} &
\textbf{ImageReward} & \textbf{PickScore} &
\textbf{IR}\,/\,\textbf{PS} \\
\midrule[0.4pt]

\multicolumn{12}{c}{\textbf{Vanilla Attention (FLUX)}} \\
\midrule[0.4pt]
Separate & \Circled[fill=method1,draw=method1,text=white]{\textbf{S}}   &  2.95 & 9.1\,/\,145.1 & 2.40 & 0.25 & 20.89 & -0.33 & 19.47 & \textbf{0.26} & \textbf{20.75} & 0.06\,/\,20.37 \\
\rowcolor{linecolor2}Fused  & \Circled[fill=method2,draw=method2,text=white]{\textbf{F}}  &  2.95 & 9.1\,/\,145.1 & 5.91 & -0.28 & 19.94 & -0.69 & 19.30 & -0.61 & 19.81 & -0.53\,/\,19.68 \\
\rowcolor{linecolor1} Hybrid & \Circled[fill=method3,draw=method3,text=white]{\textbf{H}}  &  5.90 & 18.1\,/\,145.1 & 6.86 & -0.10 & 20.36 & -0.62 & 19.24 & -0.36 & 19.98 & -0.36\,/\,19.86 \\
\rowcolor{linecolor}Tensor & \Circled[fill=method4,draw=method4,text=white]{\textbf{T}} &  2.46 & 14.1\,/\,145.1  & 7.16 & -0.01 & 20.16 & -0.44 & 19.40 & -0.38 & 19.87 & -0.28\,/\,19.81 \\

\midrule[0.4pt]
\multicolumn{12}{c}{\textbf{Cross-Intrinsic Attention~\cite{kocsis2025intrinsix}}} \\
\midrule[0.4pt]
IntrinsiX\textsuperscript{$\dagger$} & \Circled[fill=method1,draw=method1,text=white]{\textbf{S}} & 2.95 & 9.1\,/\,724.7 & 3.17 & -0.91 & 18.87 & -1.54 & 18.24 & -1.21 & 18.50 & -1.22\,/\,18.54 \\
IntrinsiX\textsuperscript{$\ddagger$} & \Circled[fill=method1,draw=method1,text=white]{\textbf{S}}  & - & - & 6.73 & -0.37 & 19.83 & -0.44 & 19.73 & - & - & -0.41\,/\,19.78 \\
\rowcolor{linecolor}IntrinsiX\textsuperscript{$\dagger$}  & \Circled[fill=method4,draw=method4,text=white]{\textbf{T}}  & 2.46 & 14.1\,/\,724.7 & 7.65 & -0.43 & 19.62 & -0.62 & 19.31 & -0.51 & 19.64 & -0.52\,/\,19.52 \\
\rowcolor{linecolor}IntrinsiX  & \Circled[fill=method4,draw=method4,text=white]{\textbf{T}}     & 2.46 & 14.1\,/\,724.7 & 7.98 & -0.05 & 20.01 & -0.48 & 19.28 & -0.30 & 19.85 & -0.28\,/\,19.71 \\

\midrule[0.4pt]
\multicolumn{12}{c}{\textbf{Query-Broadcast Attention (Ours)}} \\
\midrule[0.4pt]
Separate  & \Circled[fill=method1,draw=method1,text=white]{\textbf{S}}   & \textbf{2.16} & \textbf{6.7}\,/\,145.1 & 4.40 & -0.06 & 20.46 & -0.14 & 19.76 & -0.16 & 20.40 & -0.12\,/\,20.21 \\
\rowcolor{linecolor2}Fused    & \Circled[fill=method2,draw=method2,text=white]{\textbf{F}}     & 2.16 & 6.7\,/\,145.1 & 6.82 & 0.05 & 20.28 & -0.47 & 19.45 & -0.29 & 20.06 & -0.24\,/\,19.93 \\
\rowcolor{linecolor1} \textbf{\ours}   & \Circled[fill=method3,draw=method3,text=white]{\textbf{H}}    & 4.03 & 12.4\,/\,\textbf{145.1} & 8.21 & 0.40 & 20.32 & -0.04 & 19.74 & 0.17 & 20.31 & 0.18\,/\,20.12 \\
\rowcolor{linecolor}\textbf{\ours} & \Circled[fill=method4,draw=method4,text=white]{\textbf{T}} & 2.34  & 12.1\,/\,\textbf{145.1} & \textbf{8.30} & \textbf{0.45} & \textbf{21.02} & \textbf{0.04} & \textbf{19.88} & 0.09 & 20.66 & \textbf{0.19}\,/\,\textbf{20.52} \\

\bottomrule[0.7pt]
\end{tabular}
}

\vspace{-5pt}
\end{table*}

%% file: pic/Comparison_Results.tex
\begin{figure*}[t!]
    \begin{center}
    \includegraphics[width=\linewidth]{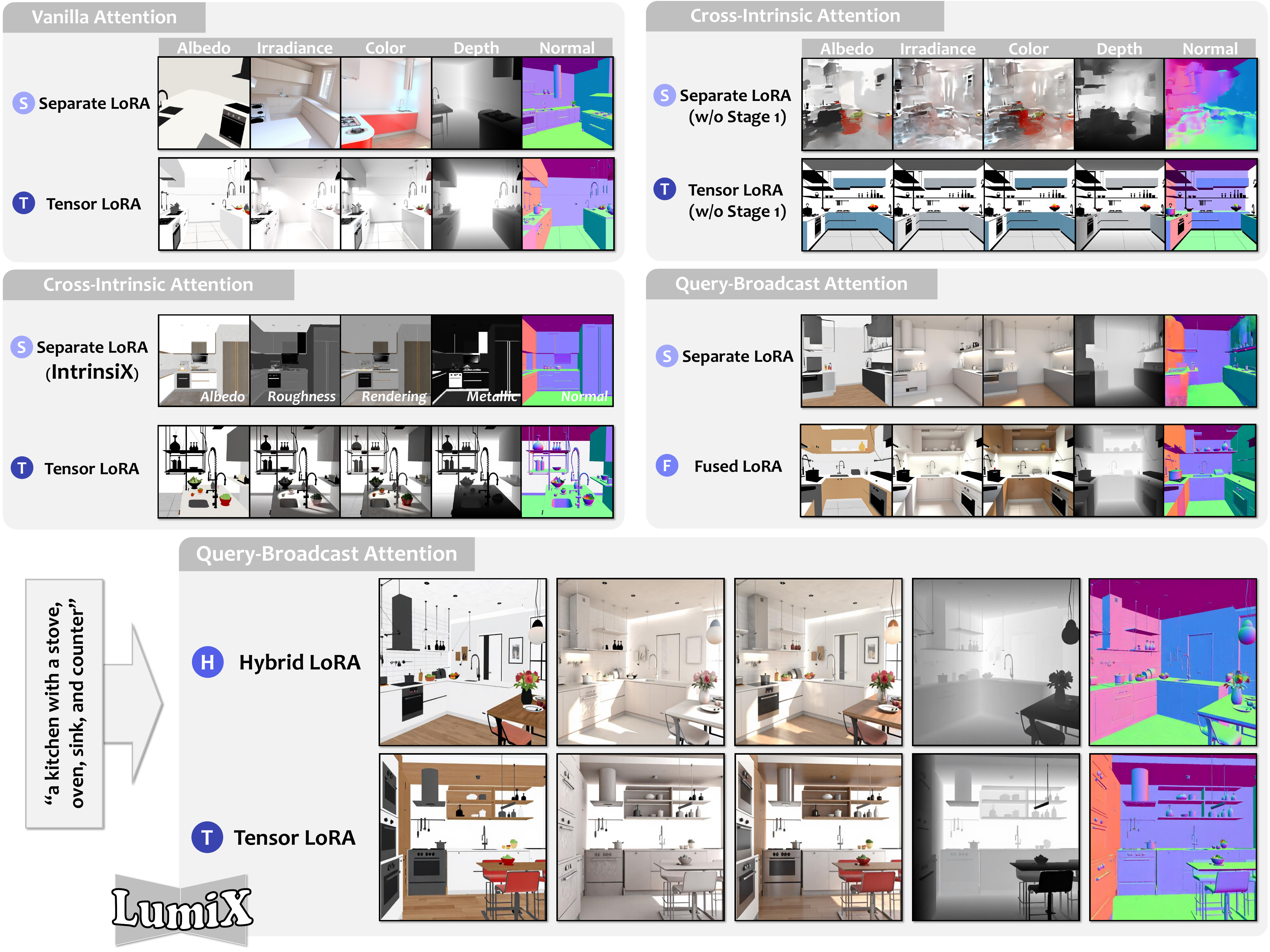}
    \vspace{-20pt}
    \caption{
\textbf{Visual comparison of Attention and LoRA designs.} 
Using vanilla attention with separate LoRA leads to the weakest alignment.  
Replacing it with our Tensor LoRA improves consistency and quality.  
IntrinsiX without its first training stage becomes unstable, while substituting Tensor LoRA alleviates collapse but still struggles to distinguish different modality characteristics.  
Our Query-Broadcast Attention combined with Hybrid or Tensor LoRA achieves the best results, producing consistent and high-quality intrinsic maps.
    }\label{fig:method_comparison_results}
    \vspace{-20pt}
    \end{center}
\end{figure*}

%% file: sec/4_experiment.tex
\section{Experiment}
\label{sec:experiment}

\subsection{Implementation Details}

\vspace{4pt}
\noindent
\textbf{Training Details.}
We fine-tune the FLUX.1-dev model~\cite{flux2023} on a subset of the Hypersim dataset~\cite{roberts2021hypersim} containing about 3K images, which is sufficient to produce stable and high-quality results.  
We use BLIP-2~\cite{li2023blip} to generate captions from color images as text inputs.  
For Tensor LoRA, we set the rank to $8$, resulting in about 133.1M trainable parameters in total.  
We use the Prodigy optimizer~\cite{mishchenko2023prodigy} with a learning rate of $1.0$ and a batch size of 16, and train the model for 10K steps on four NVIDIA A100 (80GB) GPUs, taking about 40 hours.  
All images are resized to $512{\times}512$ resolution using ratio-preserving scaling and random cropping for data augmentation, following RGB$\leftrightarrow$X~\cite{zeng2024rgb}.

\vspace{4pt}
\noindent
\textbf{Disentangled Timestep Sampling.}
Inspired by prior works~\cite{chen2024diffusion,byung2025jointdit}, we adopt a \emph{timestep-disentangled} scheme that assigns each intrinsic property an independent diffusion timestep.  
This introduces different noise levels across properties, acting as a soft mask that encourages flexible conditioning.  
Although the model is trained with text-only conditioning, it naturally supports {image-conditioned generation} at inference by keeping one property clean while denoising others.  
Combined with Query-Broadcast Attention, this ensures structural consistency between generated maps and the conditioning image, enabling intrinsic decomposition as well as diverse conditional and editing tasks.

\vspace{4pt}
\noindent
\textbf{Baselines.}
We compare \ours with IntrinsiX~\cite{kocsis2025intrinsix}, the first model targeting text-to-PBR generation. IntrinsiX produces albedo, material, and normal maps, but does not generate irradiance, depth, or full-color images. For fairness, we reproduce its stage-2 training using separate LoRA and vanilla attention, and evaluate albedo, irradiance, and color outputs.
Since IntrinsiX cannot perform intrinsic image decomposition, we additionally include RGB$\leftrightarrow$X and Colorful Shading~\cite{careaga2024colorful}, two state-of-the-art decomposition baselines trained on large-scale datasets, as references for albedo quality and cross-map consistency.
Finally, we conduct ablations on different attention and LoRA designs to analyze their effect on image quality and structural alignment.

\input{tab/Intrinsic_Decomposition_Table}

\input{pic/Text_to_Intrinsics_Results}

\input{pic/Intrinsic_Decomposition}

\vspace{4pt}
\noindent
\textbf{Evaluation Metrics.}
We use two types of quantitative evaluation.
For text-to-intrinsic generation, evaluating the quality of intrinsic maps is challenging because no ground-truth supervision is available. We therefore use human preference score models ImageReward~\cite{xu2023imagereward} and PickScore~\cite{kirstain2023pick} to assess perceptual quality. To measure cross-map consistency, we use Qwen3-VL~\cite{yang2025qwen3}, a strong vision-language evaluator. We sample 200 images from the Hypersim test set, generate captions with BLIP-2~\cite{li2023blip}, and use them as both model inputs and evaluation prompts.
For intrinsic decomposition, we follow prior work~\cite{careaga2024colorful} and perform zero-shot evaluation on ARAP~\cite{bonneel2017intrinsic}, reporting RMSE and SSIM for albedo reconstruction. To further examine generalization, we also evaluate on 50 in-the-wild photographs collected from the Internet, covering diverse indoor and outdoor scenes.

\subsection{Text-to-Intrinsic Generation}
\vspace{4pt}
\noindent
\textbf{Comparisons on the Hypersim dataset.}
\cref{tab:component_ablation,fig:method_comparison_results} compare different attention and LoRA designs for text-to-intrinsic generation.
The vanilla FLUX attention with Separate LoRA produces high-quality individual maps but lacks cross-map coherence, resulting in the lowest alignment score.  
{Fused LoRA} and {Hybrid LoRA} partially address this issue through shared updates, yet they either compromise fidelity or incur a higher parameter cost.  
In contrast, our {Tensor LoRA} achieves a better trade-off between quality, consistency, and efficiency.
We also evaluate the {Cross-Intrinsic Attention} from IntrinsiX.
Without stage-1 initialization using vanilla attention, it tends to collapse since all properties rely on globally shared $K$ and $V$.
Stage-1 pretraining improves stability, but replacing Separate LoRA with our Tensor LoRA enables robust joint training even without such initialization.
Finally, integrating our {Query-Broadcast Attention} improves all LoRA variants.
Fused LoRA achieves moderate alignment gains, while Hybrid LoRA enhances structural coherence at the cost of more parameters.
Combining Query-Broadcast Attention with Tensor LoRA yields the best overall performance.
As shown in \cref{fig:text_to_intrinsics_results}, our model produces visually coherent intrinsic maps with high perceptual quality and accurate geometric consistency.

\input{pic/X_to_X_Generation}

\subsection{Intrinsic Image Decomposition}

\vspace{4pt}
\noindent
\textbf{Albedo quality on ARAP dataset.}
Although our model is trained only with text conditioning, it can also take an image as input during inference to perform intrinsic decomposition.  
We follow the Colorful Shading~\cite{careaga2024colorful} setup and evaluate albedo quality on the ARAP dataset, as shown in \cref{tab:arap}.  
Models without diffusion priors are usually trained from scratch on large datasets and can achieve strong results.  
Diffusion-based methods typically build on Stable Diffusion U-Net architectures and are trained specifically for image decomposition.  
In contrast, our model is based on the FLUX DiT architecture and is trained without any image conditioning or ControlNet~\cite{zhang2023adding} design.  
Even so, it achieves comparable or better performance than state-of-the-art diffusion-based baselines.  
Moreover, it is the only model that can also generate intrinsic maps directly from text.

\input{tab/Albedo_Wild_Table}

\vspace{4pt}
\noindent
\textbf{Generalization to in-the-wild data.}
We evaluate our model on an {in-the-wild} dataset containing diverse real-world photographs from both indoor and outdoor scenes.
As shown in \cref{tab:albedo_wild}, our method outperforms state-of-the-art intrinsic decomposition models,
achieving the highest preference scores (ImageReward and PickScore) despite being trained on a much smaller dataset (3K vs. 900K).
Qualitative comparisons in \cref{fig:intrinsic_decomposition} further confirm that our approach generalizes well to real-world imagery.
Our generated albedo maps contain less baked-in lighting, and the predicted normals remain geometrically consistent under complex illumination.

\input{tab/Albation_Table}

\subsection{Ablation Study}
\label{sec:ablation}

\cref{tab:ablation_small} reports ablations on the Hypersim dataset.
Fine-tuning the Query projection $W_Q$ in our Query-Broadcast Attention increases parameters and FLOPs but degrades intrinsic quality,
as it disrupts the diffusion prior and confuses learning objectives.
Varying the rank $R_1=R_2=R$, we find $R=8$ gives the best balance of quality and efficiency,
while $R=4$ remains competitive with minimal parameters.

%% file: tab/Intrinsic_Decomposition_Table.tex
\begin{table}[t]
\renewcommand{\arraystretch}{1.1}
\centering
\caption{
\textbf{Zero-shot albedo evaluation on the synthetic ARAP Dataset}~\cite{bonneel2017intrinsic}.
We evaluate intrinsic decomposition quality using RMSE and SSIM.
T2I indicates whether each method supports text-to-image generation.
$^*$ denotes non–zero-shot methods.
Our method is the only one that supports text-to-intrinsic generation.
}
\label{tab:arap}

\small
\setlength{\tabcolsep}{5.5pt}
\resizebox{\columnwidth}{!}{
\begin{tabular}{lcccc}
\toprule[0.7pt]
\textbf{Method} & \textbf{T2I} & \textbf{Base} & \textbf{RMSE}$\downarrow$ & \textbf{SSIM}$\uparrow$ \\
\midrule[0.4pt]
\multicolumn{5}{c}{\textit{Without Diffusion Prior}} \\
\midrule[0.4pt]
Chromaticity & \ding{55} & - & 0.193 & 0.710 \\
Constant Shading & \ding{55} & - & 0.264 & 0.693 \\
\citet{lettry2018unsupervised} & \ding{55} & - & 0.193 & 0.732 \\
NIID-Net~\cite{luo2020niid}$^*$ & \ding{55} & - & \textbf{0.129} & 0.788 \\
\citet{zhu2022learning} & \ding{55} & - & 0.184 & 0.729 \\
Ordinal Shading~\cite{careaga2023intrinsic} & \ding{55} & - & 0.162 & 0.751 \\
Colorful Shading~\cite{careaga2024colorful} & \ding{55} & - & {0.149} & \textbf{0.796} \\
\midrule[0.4pt]
\multicolumn{5}{c}{\textit{With Diffusion Prior}} \\
\midrule[0.4pt]
IntrinsicAnything~\cite{chen2024intrinsicanything} & \ding{55} & SD-CLIP$_I$ & 0.171 & 0.692 \\
IID~\cite{kocsis2024intrinsic} & \ding{55} & SD2-Depth & \textbf{0.160} & 0.738 \\
RGB$\leftrightarrow$X~\cite{zeng2024rgb} & \ding{55} & SD2.1 & 0.238 & 0.564 \\
\rowcolor{linecolor}\textbf{\ours (Ours)} & \color{newgreen}{\textbf{\ding{51}}} & FLUX & 0.165 & \textbf{0.753} \\
\bottomrule[0.7pt]
\end{tabular}
}
\end{table}

%% file: pic/Text_to_Intrinsics_Results.tex
\begin{figure*}[t!]
    \begin{center}
    \vspace{-4pt}
    \includegraphics[width=\linewidth]{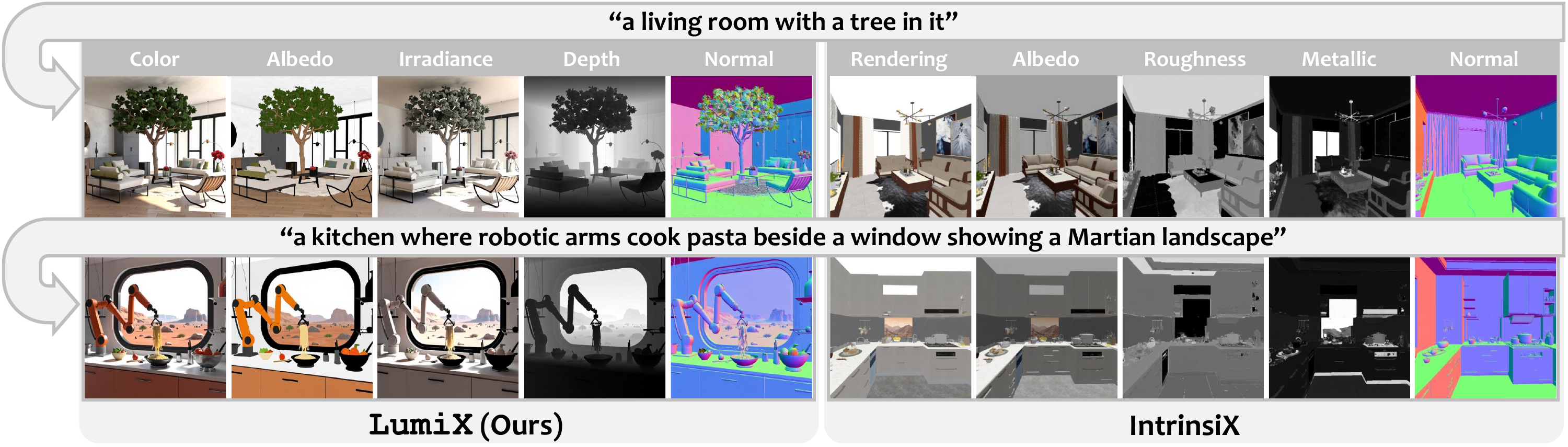}
    \vspace{-20pt}
\caption{
\textbf{Text-to-Intrinsic Generation Comparison.} 
Both models are built upon FLUX. 
While IntrinsiX tends to overfit to specific indoor scenes, our method preserves FLUX's strong prior and produces consistent, high-quality intrinsic maps even for out-of-domain prompts.
    }\label{fig:text_to_intrinsics_results}
    \vspace{-10pt}
    \end{center}
\end{figure*}

%% file: pic/Intrinsic_Decomposition.tex
\begin{figure*}[t!]
    \begin{center}
    \vspace{-4pt}
    \includegraphics[width=\linewidth]{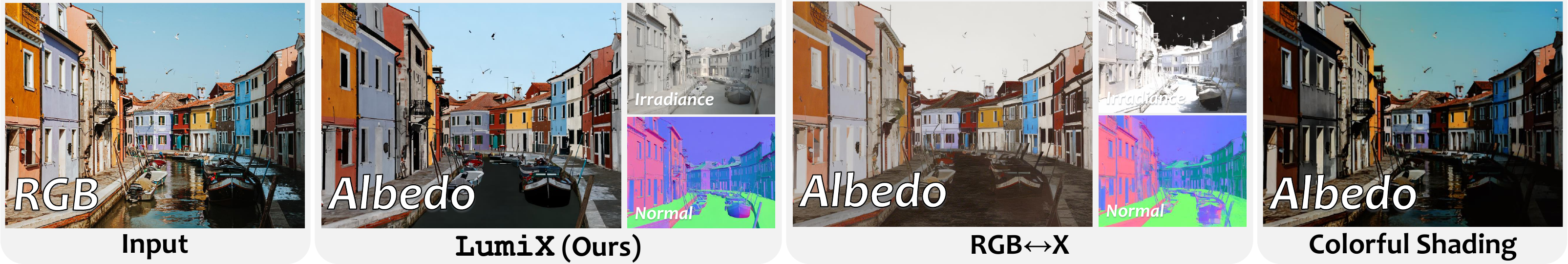}
    \vspace{-20pt}
\caption{
\textbf{Intrinsic Decomposition Comparison.}  
Our method performs intrinsic decomposition on {in-the-wild} data, producing albedo maps with less embedded lighting and generating consistent, high-quality intrinsic maps across all properties.
    }\label{fig:intrinsic_decomposition}
    \vspace{-20pt}
    \end{center}
\end{figure*}

%% file: pic/X_to_X_Generation.tex
\begin{figure*}[t!]
    \begin{center}
    \vspace{-4pt}
    \includegraphics[width=\linewidth]{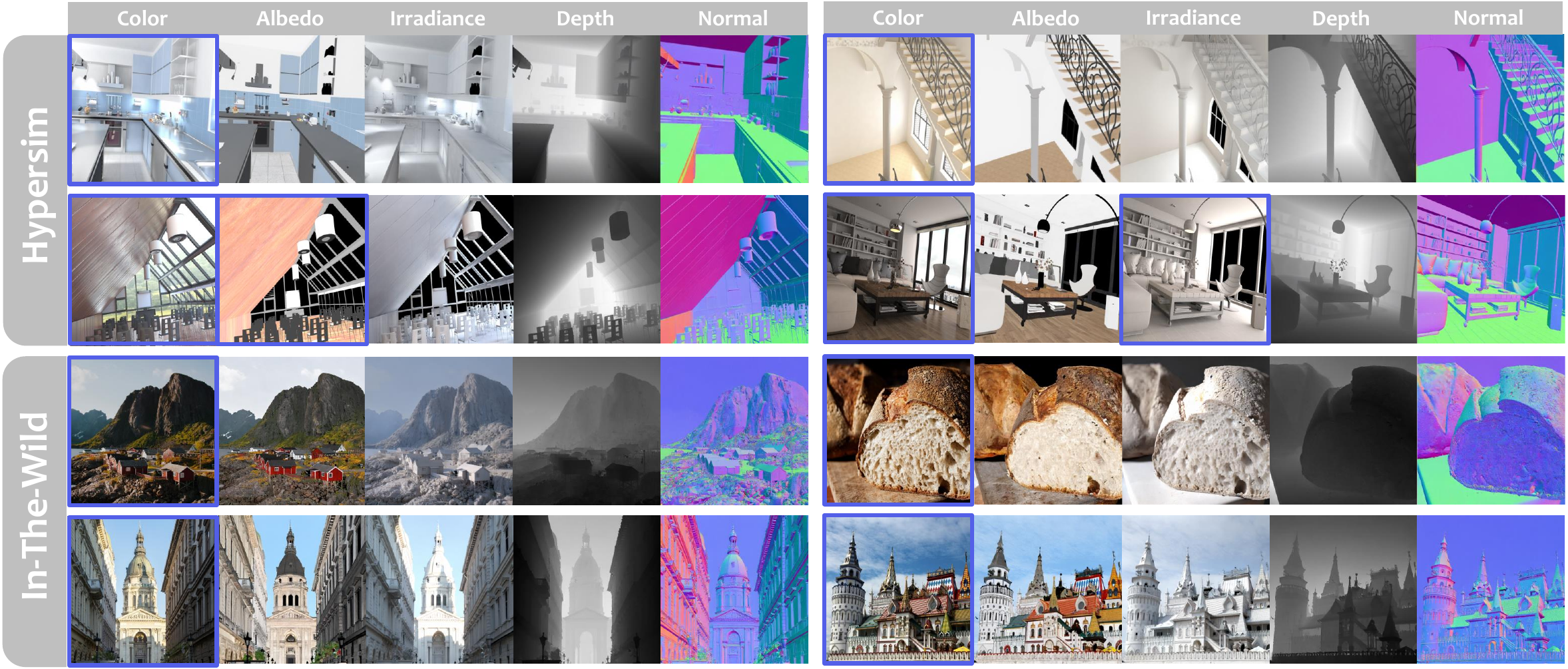}
    \vspace{-20pt}
    \caption{
\textbf{Image-Conditioned Intrinsic Decomposition.}  
Given a color image, our model performs intrinsic decomposition on Hypersim and {in-the-wild} data, 
and can further take albedo or irradiance as additional conditions. Conditioned images are shown with \fcolorbox{method4}{white}{blue boxes}.
    }\label{fig:x_to_x_results}
    \vspace{-20pt}
    \end{center}
\end{figure*}

%% file: tab/Albedo_Wild_Table.tex
\begin{table}[t]
\renewcommand{\arraystretch}{1.1}
\centering
\caption{
\textbf{Zero-shot albedo evaluation on our {in-the-wild} dataset}.  
We evaluate intrinsic decomposition quality using ImageReward and PickScore across different variants of our method.
}
\label{tab:albedo_wild}
\vspace{-2pt}
\small
\setlength{\tabcolsep}{6pt}
\resizebox{0.85\columnwidth}{!}{ %
\begin{tabular}{lcc}
\toprule[0.7pt]
\textbf{Method} & \textbf{ImageReward$\uparrow$} & \textbf{PickScore$\uparrow$} \\
\midrule[0.4pt]
RGB$\leftrightarrow$X~\cite{zeng2024rgb} & -0.20 & 20.01 \\
Colorful Shading~\cite{careaga2024colorful} & 0.06 & 20.03 \\
\rowcolor{linecolor}\textbf{\ours (Ours)} & \textbf{0.14} & \textbf{20.16} \\
\bottomrule[0.7pt]
\end{tabular}
}
\vspace{-10pt}
\end{table}

%% file: tab/Albation_Table.tex
\begin{table}[t]
\renewcommand{\arraystretch}{1.1}
\centering
\caption{
\textbf{Ablation of different variants of our method.}
We report the averaged ImageReward and PickScore.
\#P: trainable parameters (M) per attention block;  
\#F: FLOPs (G) of LoRA per block.
}
\label{tab:ablation_small}
\vspace{-2pt}
\small
\setlength{\tabcolsep}{4pt} %
\resizebox{\columnwidth}{!}{ %
\begin{tabular}{lccccc}
\toprule[0.7pt]
\multirow{2}{*}{\textbf{Method}} &
\multirow{2}{*}{\textbf{\#P$\downarrow$}} &
\multirow{2}{*}{\textbf{\#F$\downarrow$}} &
\multirow{2}{*}{\textbf{Align.$\uparrow$}} &
\multicolumn{2}{c}{\textbf{Average}} \\
\cmidrule(lr){5-6}
& & & & \textbf{ImageReward}$\uparrow$ & \textbf{PickScore}$\uparrow$ \\
\midrule[0.4pt]
\rowcolor{linecolor}\textbf{\ours} & {2.34} & 12.1 & \textbf{8.30} & \textbf{0.19} & \textbf{20.52} \\
+ Tune $W_Q$ & 2.46 & 14.1 & 7.14 & -0.09 & 20.04 \\
\midrule[0.4pt]
$R = 4$ & \textbf{0.68} & \textbf{4.7} & 7.86 & -0.18 & 19.79 \\
$R = 8$ & {2.34} & 12.1 & {8.30} & {0.19} & {20.52} \\
$R = 12$ & 4.98 & 22.5 & 8.10 & 0.14 & 20.29 \\
\bottomrule[0.7pt]
\end{tabular}
}
\vspace{-10pt}
\end{table}

%% file: sec/5_conclusion.tex
\section{Conclusion}
\label{sec:conclusion}
We presented \ours, a structured diffusion framework for text-to-intrinsic generation. By combining Query-Broadcast Attention, which enforces pixel-level alignment across intrinsic maps, with Tensor LoRA, which provides lightweight yet expressive parameter adaptation, \ours produces coherent intrinsic images with strong structural consistency and high visual fidelity. Our experiments show that stable multi-map generation arises not only from scaling up supervision, but also from structured parameter sharing. 
We plan to scale \ours to a broader set of intrinsic properties and larger datasets in the future, moving toward models that achieve unified understanding and generation of scene geometry, materials, and illumination.

%% file: sec/X_suppl.tex
\clearpage
\setcounter{page}{1}
\maketitlesupplementary

\renewcommand\thesection{\Alph{section}}


\section*{A. Method Details}

\subsection*{A.1. Additional Details on Query Broadcast Attention}

\paragraph{Pseudocode}
We provide minimal pseudocode for the Query Broadcast Attention forward pass in~\cref{alg:qba}.

\begin{algorithm}[h]
\caption{Forward pass of Query Broadcast Attention}
\label{alg:qba}
\begin{algorithmic}[1]
\Require $x \in \mathbb{R}^{M \times L \times D}$  \Comment{$M$ intrinsic properties, $L$ tokens per property, feature dimension $D$}
\Require $\text{attn}$ (attention block with $H$ heads)
\Require $\text{TensorLoRA}^{K}$, $\text{TensorLoRA}^{V}$ \Comment{Tensor LoRA applied to $K$/$V$ projections}
\Require $\text{pe}$ \Comment{positional encoding}

\State $q, k, v \gets \text{attn.qkv}(x)$  
\Comment{$[M, H, L, D_{\text{head}}]$}

\State $q_{\text{shared}} \gets q[\text{color}]$  
\Comment{$[1, H, L, D_{\text{head}}]$}

\State $q_{\text{shared}} \gets \text{expand}(q_{\text{shared}}, M)$  
\Comment{$[M, H, L, D_{\text{head}}]$}

\State $\Delta k \gets \text{TensorLoRA}^{K}(x)$  \Comment{$[M, L, D]$}
\State $\Delta v \gets \text{TensorLoRA}^{V}(x)$  \Comment{$[M, L, D]$}

\State $\Delta k \gets \text{reshape\_to\_heads}(\Delta k)$  \Comment{$[M, H, L, D_{\text{head}}]$}
\State $\Delta v \gets \text{reshape\_to\_heads}(\Delta v)$  \Comment{$[M, H, L, D_{\text{head}}]$}

\State $k_{\text{mod}} \gets k + \Delta k$
\State $v_{\text{mod}} \gets v + \Delta v$

\State $y \gets \text{attention}(q_{\text{shared}}, k_{\text{mod}}, v_{\text{mod}}, \text{pe})$

\State \Return $y$
\end{algorithmic}
\end{algorithm}

\subsection*{A.2. Additional Details on Tensor LoRA}

\paragraph{Fused Contraction}
The main paper presents Tensor LoRA using three tensor contractions:
\begin{verbatim}
# N d R1 R2, M d -> N M R1 R2
Ch   = einsum('ndrs, md -> nmrs', C, h)
# N M R2, N M R1 R2 -> N R1
BCh  = einsum('nms, nmrs -> nr', B, Ch)
# N d R1, N R1 -> N d
ABCh = einsum('ndr, nr -> nd', A, BCh)
\end{verbatim}
These operations follow directly from the element-wise form in
Eq.~\ref{equ:tensor_lora}, where the rank indices
$(\alpha_1,\alpha_2)$ are contracted across the three components
$A$, $B$, and $C$.

\paragraph{Single-Step Equivalent}
Because all intermediate indices are internal and do not appear in the
final output, the entire computation can be expressed as a single
contraction:
\begin{equation}
\label{eq:fused_supp}
\small
\Delta[i,o] =
\sum_{j,l,\alpha_1,\alpha_2}
C[i,l,\alpha_1,\alpha_2]\,
h[j,l]\,
B[i,j,\alpha_2]\,
A[i,o,\alpha_1].
\end{equation}

This form can be implemented using a single \textsf{einsum} call:
\begin{verbatim}
ABCh = einsum('ndrs, md, nms, ndr -> nd',
              C, h, B, A)
\end{verbatim}
which avoids constructing the intermediate tensors in the three-step
version and improves memory efficiency.

\section*{B. Implementation Details}
\paragraph{Additional Training Details}
All intrinsic properties are processed together by concatenating them
along the batch dimension.
All weights of FLUX.1-dev\footnote{\url{https://github.com/black-forest-labs/flux}}
remain frozen during training.
The only trainable parameters are the Tensor LoRA updates inserted into
the key and value projections of each attention block, while all original
query, key, and value weights are kept fixed.
The flow-matching loss is computed independently for each intrinsic
property and then averaged.
For data preparation, we follow the official Hypersim\footnote{\url{https://github.com/apple/ml-hypersim}} pipeline and apply
their tone mapping and gamma correction to convert HDR color,
irradiance, and albedo images into LDR inputs.

\paragraph{Evaluation Metrics}
Evaluating intrinsic images such as albedo is inherently challenging due
to their ill-posed nature. To assess perceptual quality, we use two
state-of-the-art human preference models, ImageReward%
\footnote{\url{https://github.com/zai-org/ImageReward}} and PickScore%
\footnote{\url{https://github.com/yuvalkirstain/PickScore}},
which we find to correlate well with visual realism in our setting.
To measure cross-property consistency, we rely on Qwen3-VL%
\footnote{\url{https://huggingface.co/Qwen/Qwen3-VL-4B-Instruct}},
a strong vision–language model capable of fine-grained visual reasoning.
For each pair of intrinsic maps (\eg, color and albedo), the model is
asked to evaluate their alignment. It assigns five global scores in
$[0,1]$ for scene identity, structural layout, object arrangement,
material–region consistency, and reflectance correspondence, summed to a
global score in $[0,5]$.  
If the global score is at least $2.0$, it further assigns five local
scores in $[0,1]$ for edge precision, geometric detail, surface
boundaries, shadow/illumination separation, and small-object/material
consistency, yielding a local score in $[0,5]$.  
The final alignment score is the total in the interval $[0,10]$.

\section*{C. Additional Results}
We include additional examples from our quantitative evaluation in~\cref{fig:supp_intrinsic_decomp_p1,fig:supp_intrinsic_decomp_p2,fig:supp_intrinsic_decomp_p3,fig:supp_intrinsic_decomp_p4,fig:supp_intrinsic_decomp_p5,fig:supp_text_gen_p1}.
\input{suppl/supp_text_gen}

\input{suppl/supp_intrinsic_decomp_p1}

\input{suppl/supp_intrinsic_decomp_p2}

\input{suppl/supp_intrinsic_decomp_p3}

\input{suppl/supp_intrinsic_decomp_p4}

\input{suppl/supp_intrinsic_decomp_p5}

%% file: suppl/supp_text_gen.tex
\begin{figure*}[t]
    \begin{center}
    \includegraphics[width=\linewidth]{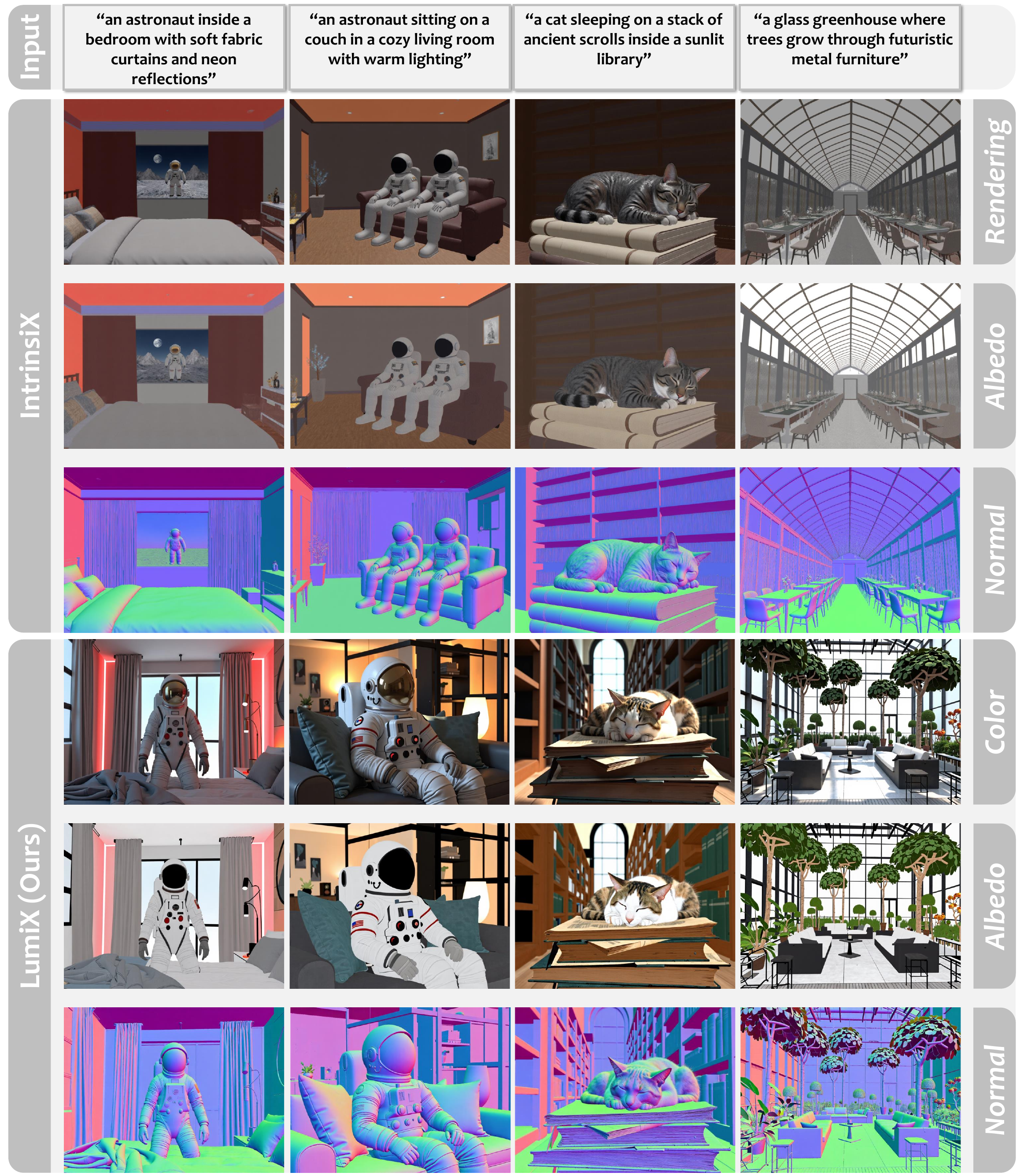}
\caption{\textbf{More Text-to-Intrinsic Generation Results}. 
Compared with IntrinsiX~\cite{kocsis2025intrinsix}, our method yields higher-quality images and consistent intrinsic maps.
}\label{fig:supp_text_gen_p1}
    \vspace{-20pt}
    \end{center}
\end{figure*}

%% file: suppl/supp_intrinsic_decomp_p1.tex
\begin{figure*}[t]
    \begin{center}
    \includegraphics[width=\linewidth]{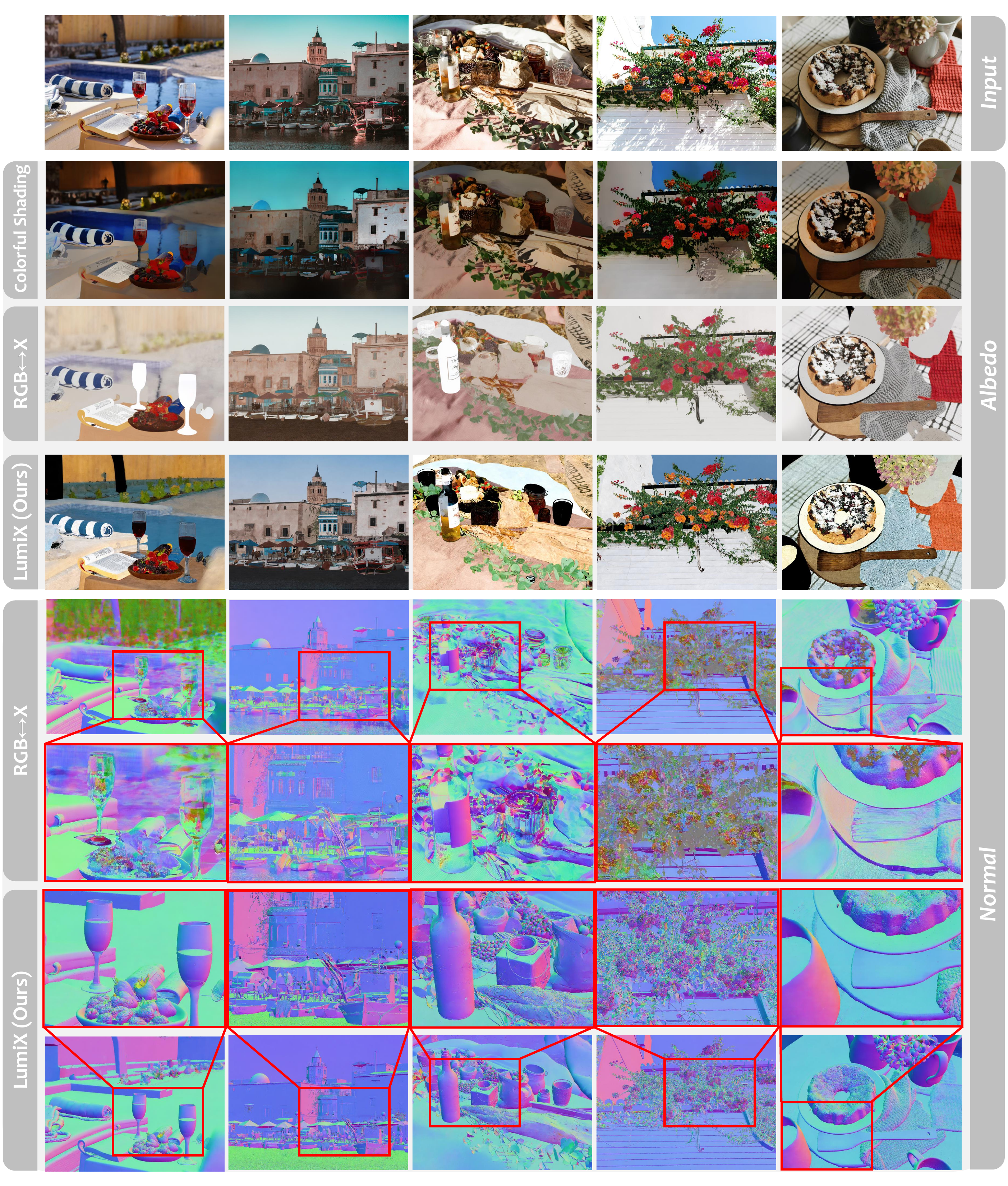}
    \caption{\textbf{More Intrinsic Decomposition Results}. 
Our method produces cleaner albedo and more accurate, consistent normal maps compared with prior baselines.
    }\label{fig:supp_intrinsic_decomp_p1}
    \vspace{-20pt}
    \end{center}
\end{figure*}

%% file: suppl/supp_intrinsic_decomp_p2.tex
\begin{figure*}[t]
    \begin{center}
    \includegraphics[width=\linewidth]{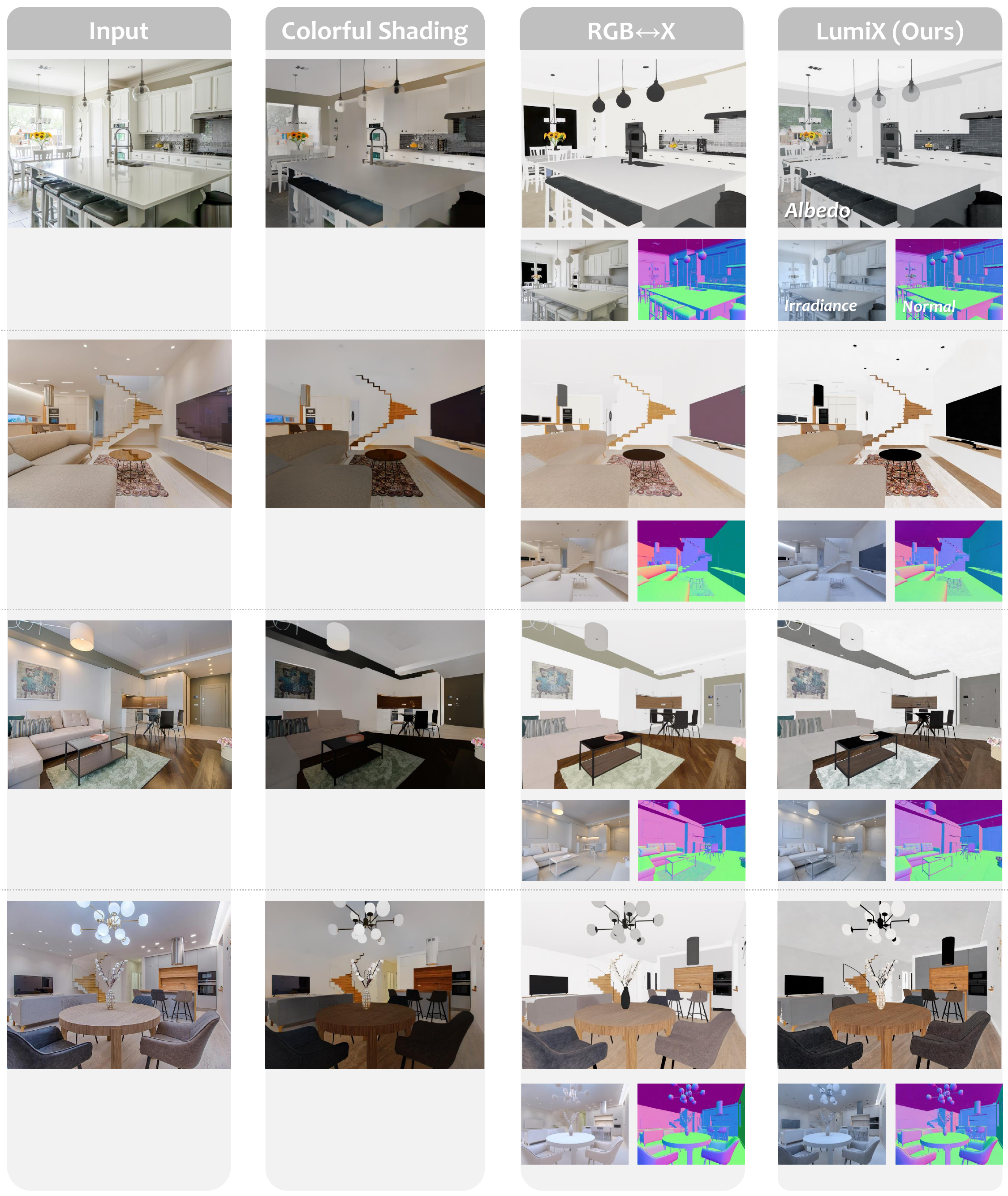}
    \caption{\textbf{More Intrinsic Decomposition Results on Indoor Scenes}. We show the albedo, irradiance, and normal predictions for RGB$\leftrightarrow$X~\cite{zeng2024rgb} and our method. Both methods produce high-quality albedo and irradiance, while ours yields more accurate and consistent normal maps.
    }\label{fig:supp_intrinsic_decomp_p2}
    \vspace{-20pt}
    \end{center}
\end{figure*}

%% file: suppl/supp_intrinsic_decomp_p3.tex
\begin{figure*}[t]
    \begin{center}
    \includegraphics[width=\linewidth]{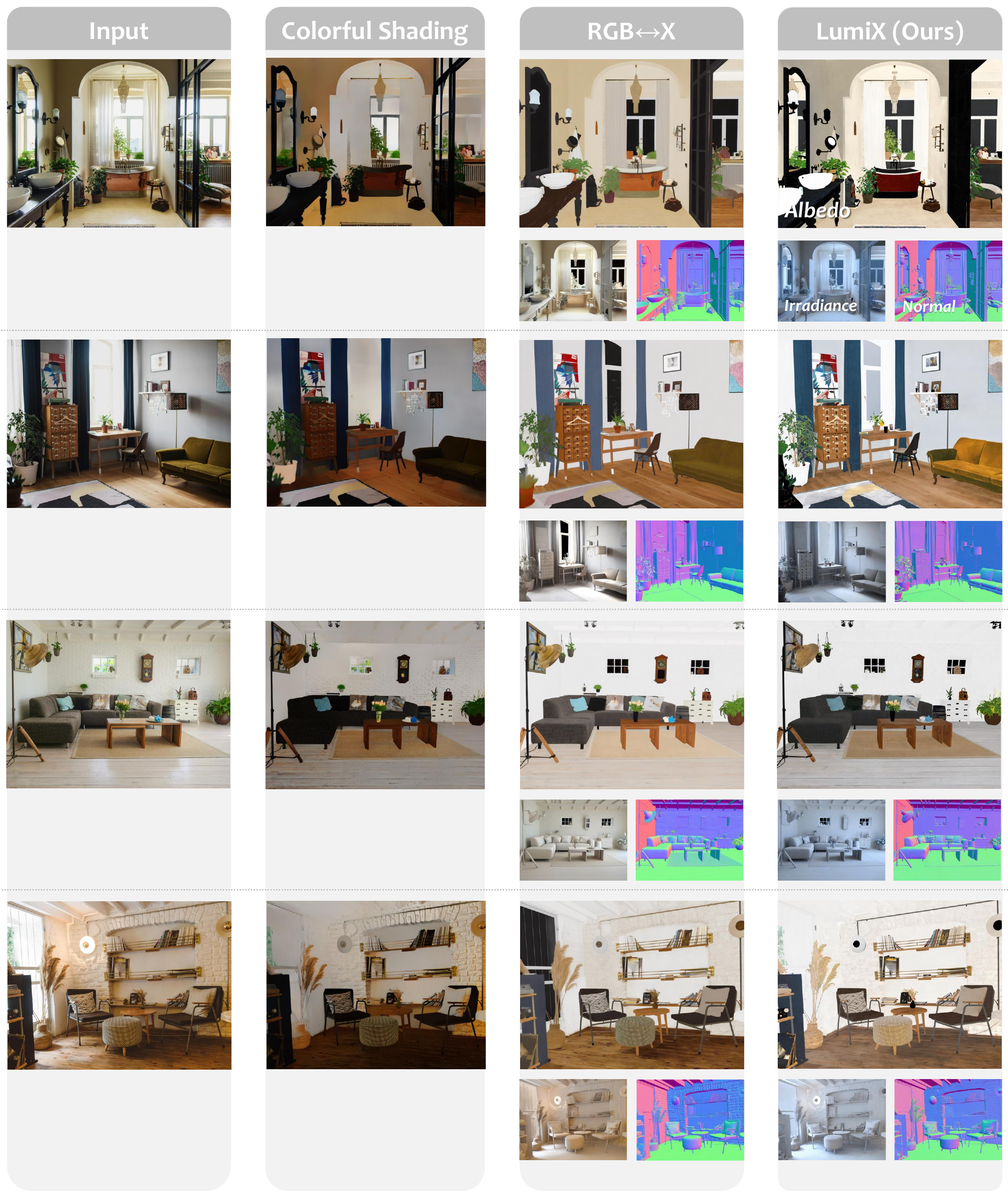}
    \caption{\textbf{More Intrinsic Decomposition Results on Indoor Scenes}. We visualize albedo, irradiance, and normal maps from RGB$\leftrightarrow$X~\cite{zeng2024rgb} and our method.  Both achieve strong albedo and irradiance performance, but ours attains higher overall quality.
}\label{fig:supp_intrinsic_decomp_p3}
    \vspace{-20pt}
    \end{center}
\end{figure*}

%% file: suppl/supp_intrinsic_decomp_p4.tex
\begin{figure*}[t]
    \begin{center}
    \includegraphics[width=\linewidth]{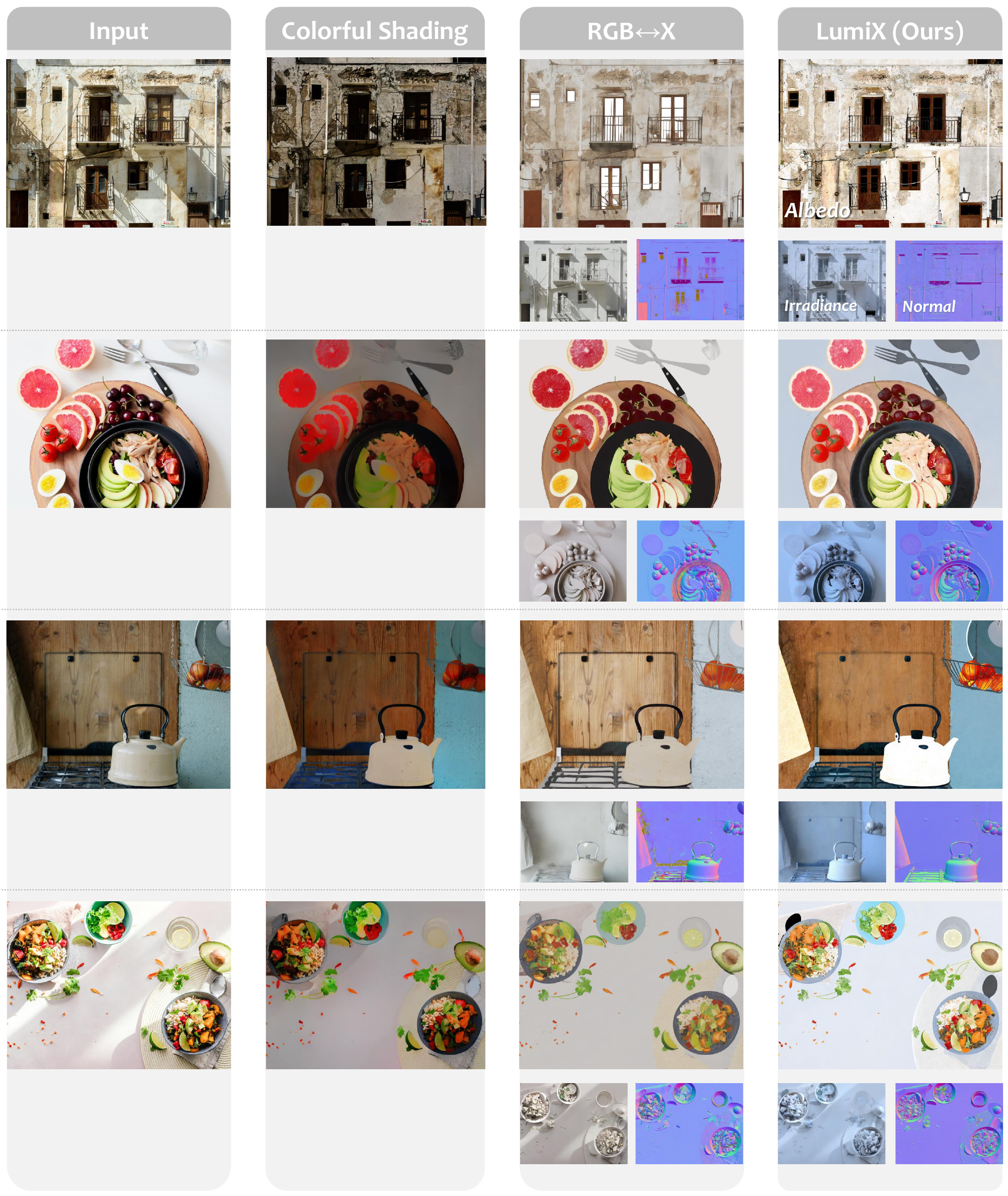}
    \caption{\textbf{More Intrinsic Decomposition Results on Out-of-Domain Data}. 
Our method produces high-quality and consistent intrinsic maps even outside indoor scenes, with cleaner albedo (less baked-in lighting) and more accurate normals.}
\label{fig:supp_intrinsic_decomp_p4}
    \vspace{-20pt}
    \end{center}
\end{figure*}

%% file: suppl/supp_intrinsic_decomp_p5.tex
\begin{figure*}[t]
    \begin{center}
    \includegraphics[width=\linewidth]{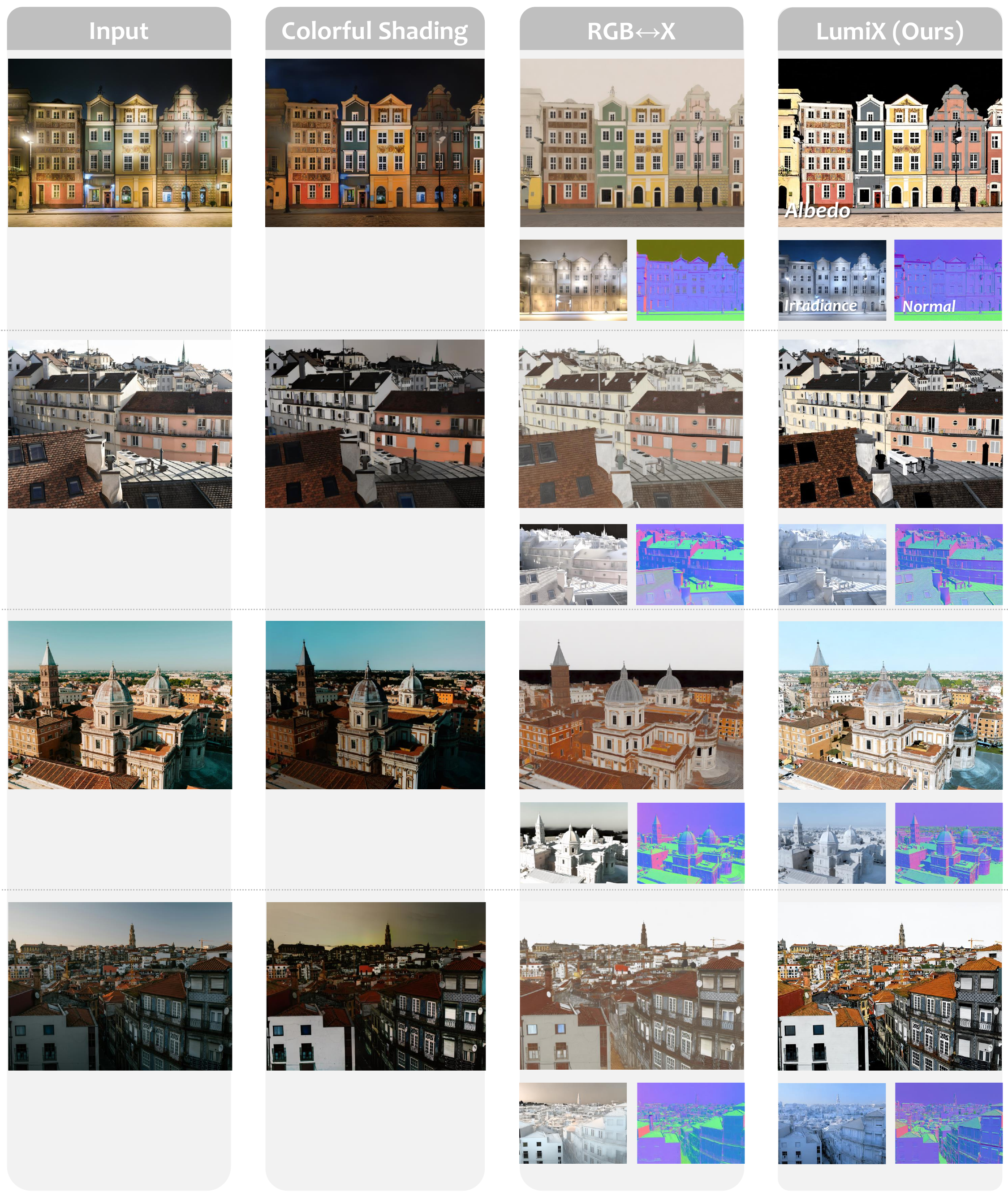}
    \caption{\textbf{More Intrinsic Decomposition Results on Out-of-Domain Data}. 
We visualize albedo, irradiance, and normal maps on clearly out-of-domain images.
Despite being trained on only 3K examples (far fewer than the 190K images used by RGB$\leftrightarrow$X~\cite{zeng2024rgb}), our method still produces high-quality and consistent intrinsic maps, with cleaner albedo and more accurate normal maps.}
\label{fig:supp_intrinsic_decomp_p5}
    \vspace{-20pt}
    \end{center}
\end{figure*}